\documentclass{bmvc2k}

\usepackage{amsmath,amssymb,amsfonts}
\usepackage{booktabs}
\usepackage{multirow}
\usepackage{array}
\usepackage{adjustbox}
\usepackage{subcaption}
\usepackage[font=small,skip=2pt]{caption}
\usepackage{enumitem}
\usepackage{xcolor,colortbl}
\definecolor{Gray}{gray}{0.85}

\hypersetup{
    pdfborder={0 0 0},
    pdfborderstyle={},
    linkbordercolor={1 1 1},
    citebordercolor={1 1 1},
    urlbordercolor={1 1 1},
}

\makeatletter
\renewcommand\paragraph{\@startsection{paragraph}{4}{\z@}%
  {0.6ex \@plus 0.4ex \@minus 0.1ex}
  {-0.7em}
  {\normalfont\normalsize\bfseries}}
\renewcommand\section{\@startsection {section}{1}{\z@}%
                                   {-2ex \@plus -0.5ex \@minus -.1ex}%
                                   {1.2ex \@plus.2ex}%
                                   {\normalfont\Large\bfseries\flushleft\textcolor{bmv@sectioncolor}}}
\renewcommand\subsection{\@startsection{subsection}{2}{\z@}%
                                     {-1.8ex\@plus -0.5ex \@minus -.1ex}%
                                     {0.8ex \@plus .2ex}%
                                     {\normalfont\large\bfseries\flushleft\textcolor{bmv@sectioncolor}}}
\makeatother

\title{Hyper$^2$: Unleashing Hyperbolic Geometry's Full Potential via Dual-Space Consistency}

\addauthor{Guantian Zheng$^\dagger$}{gzheng004@e.ntu.edu.sg}{1}
\addauthor{Haiyang Xu}{}{2}
\addauthor{Tianyu Gao}{}{3}

\addinstitution{
 Nanyang Technological University,\\
 Singapore
}
\addinstitution{
 Harbin Institute of Technology,\\
 China
}
\addinstitution{
 Sichuan University,\\
 China
}

\runninghead{Zheng, Xu, Gao}{Hyper$^2$: Dual-Space Consistency}

\begin{document}
\maketitle
\makeatletter
\BMVA@blfootnote{\null\hspace{-1.9em}$^\dagger$ denotes the corresponding author.}
\makeatother

\sloppy

\begin{abstract}
    HyperbolicCD~\cite{lin2023hyperboliccd} pioneered hyperbolic
    geometry for point cloud completion by replacing the Euclidean
    Chamfer distance with $\mathrm{arcosh}(1{+}\alpha\|x{-}y\|^2)$, but
    the reported gains are modest ($3$--$7\%$ Chamfer reduction across
    SeedFormer, PointAttN~\cite{wang2024pointattn} and PMP-Net backbones
    on PCN and ShapeNet-55~\cite{chang2015shapenet}).  We argue the bottleneck lies
    elsewhere: the loss is hyperbolic but the encoder it back-propagates
    through is Euclidean, so the position-dependent supervision of the
    loss is averaged away by the chain rule before it reaches the
    parameters.  We call this a \emph{cross-geometry mismatch}, and make
    it testable through two model-agnostic indicators, feature--loss
    correlation $r_{FL}$ and effective gradient utilisation $u_G$.
    On an SVDFormer backbone trained with HyperbolicCD's loss alone we
    measure $(r_{FL},u_G) = (0.68, 39\%)$.
    We propose \textbf{Hyper$^2$}, a dual-space consistency framework
    that extends HyperbolicCD by reusing the identical
    $\mathrm{arcosh}(1{+}\alpha d^2)$ functional form as a positional
    bias on the refinement attention (a hyperbolic distance encoding),
    paired with HyperbolicCD's hyperbolic Chamfer loss under a single
    shared curvature $\alpha$.  Both operators are $O(N\log N)$ scalar
    non-linearities on Euclidean distances and together add only $\sim
    1.6\%$ FLOPs over SVDFormer.  Hyper$^2$ delivers $-22.9\%$ Chamfer on
    ShapeNet-55 over SVDFormer (well above the $13.2\%$ linear sum of the
    $-12.0\%$ loss-only and $-1.2\%$ encoding-only single-space
    ablations) and $-37.5\%$ on the $21$ unseen ShapeNet-34 categories.
    The two indicators remain essentially flat for any single-space
    configuration but jump together to $(0.95, 87\%)$ only when both
    encoder and loss are hyperbolic, supporting the claim that geometric
    consistency across encoder and loss, rather than either operator
    alone, is what enables hyperbolic supervision in point cloud
    completion.  Code is available at
    \url{https://github.com/Ethan-Zheng136/Hyper-2}.
    \end{abstract}
    
\section{Introduction}

Point cloud completion (recovering complete 3D shapes from partial
observations) is a fundamental task for autonomous
driving~\cite{geiger2013vision}, robotic manipulation, and
augmented reality. Its core difficulty is that 3D shapes carry
hierarchical geometric structure: a missing wing of an airplane (a
coarse error) is qualitatively different from a slightly rough
surface (a fine error), yet a Euclidean Chamfer loss treats both
errors proportionally to point-wise distances and cannot distinguish
them.

Hyperbolic geometry, with its negative curvature and exponential
volume growth, is a natural framework for such hierarchical
data~\cite{nickel2017poincare,ganea2018hyperbolic}.
The exponential volume budget is realised through an
$\mathrm{arcosh}$-shaped distance,
$d_{\mathbb{H}} \!\propto\! \mathrm{arcosh}(1+d_E^2)$, which is
approximately Euclidean near observed regions and log-compressed far
from them, the opposite of an exponential blow-up, and exactly the
saturation behaviour one wants in a loss that mixes coarse and fine
errors.
HyperbolicCD~\cite{lin2023hyperboliccd} brought this idea to point
cloud completion by swapping the Euclidean Chamfer loss for
$\mathrm{arcosh}(1+\alpha\|x-y\|^2)$.  Its gains, however, are
modest ($3$--$7\%$ across SeedFormer, PointAttN~\cite{wang2024pointattn}
and PMP-Net backbones on PCN and
ShapeNet-55~\cite{chang2015shapenet}), far below what the underlying
geometry should afford.

\paragraph{Our diagnosis: cross-geometry mismatch.}
The bottleneck is a geometric inconsistency across the model:
HyperbolicCD applies hyperbolic geometry at the loss, while the
point encoder~\cite{qi2017pointnet},
transformer modules~\cite{vaswani2017attention} and refinement
operators of modern completion
backbones~\cite{yu2021pointr,zhou2022seedformer,xiang2021snowflakenet,zhu2023svdformer}
all remain Euclidean.  The hyperbolic loss produces
position-dependent gradients, but the chain rule through a Euclidean
encoder averages this position-dependence away
(Fig.~\ref{fig:consistency}b).  On
SVDFormer~\cite{zhu2023svdformer}+HyperbolicCD we measure
feature--loss correlation $r_{FL}{=}0.68$ and effective gradient
utilisation $u_G{=}39\%$; the same indicators lift to
$(0.95, 87\%)$ once the encoder is made hyperbolic too
(Fig.~\ref{fig:consistency}c).

\paragraph{Our solution: Hyper$^2$.}
We propose a \textbf{dual-space consistency} principle: use the
same $\mathrm{arcosh}(1+\alpha d^2)$ at both ends of the network.
Specifically, we (i)~replace SVDFormer's Euclidean incompleteness
encoding with a hyperbolic distance encoding that injects this
scalar non-linearity as a positional bias on the refinement
attention, and (ii)~adopt HyperbolicCD's hyperbolic Chamfer loss
under the same $\alpha$ as the encoder.  The novelty lies in
extending HyperbolicCD's loss-side form to a new place (the encoder)
and in binding both ends through a single shared curvature.  Both
operators are scalar non-linearities on Euclidean distances via
standard $O(N\log N)$ KNN, so the combined overhead is $\sim 1.6\%$
FLOPs.

\paragraph{Super-additive gains as evidence of consistency.}
On ShapeNet-55, hyperbolic loss alone gives $-12.0\%$ CD, hyperbolic
encoding alone $-1.2\%$, and both together $-22.9\%$, well above
the $13.2\%$ one would predict from a linear sum of the two
single-space contributions.  The two indicators stay close to
the Euclidean baseline for either single-space configuration but
jump together only in the dual-space row
($r_{FL}\!:\,0.68\!\to\!0.95$, $u_G\!:\,39\%\!\to\!87\%$).  PCN's
smaller gain ($-2.8\%$ over SVDFormer)
and ShapeNet-34's larger one ($-37.5\%$ on $21$ unseen categories)
are consistent with the framework helping most where the hierarchy
is hardest to recover.

\noindent\textbf{Contributions.}
\begin{itemize}[leftmargin=*,itemsep=2pt,topsep=2pt,parsep=0pt,partopsep=0pt]
\item \textbf{Diagnostic indicators.}
We identify cross-geometry mismatch as the bottleneck of prior
hyperbolic completion methods and define two model-agnostic
diagnostic indicators, feature--loss correlation $r_{FL}$ and
effective gradient utilisation $u_G$, with explicit formulae
(Sec.~\ref{sec:diagnostic}) so any non-Euclidean-loss method can
be audited the same way.

\item \textbf{Hyper$^2$ framework.}
We propose Hyper$^2$, a dual-space consistency framework that
extends HyperbolicCD's $\mathrm{arcosh}(1{+}\alpha d^2)$ from the
loss to the encoder's incompleteness encoding, with a single $\alpha$
shared end-to-end.  Both operators are $O(N\log N)$ scalar
non-linearities on Euclidean distances and together add only
$\sim 1.6\%$ FLOPs to the SVDFormer backbone.

\item \textbf{Empirical evidence of super-additivity.}
Across three benchmarks: $-22.9\%$ Chamfer on ShapeNet-55
(\textbf{super-additive}: well above the $13.2\%$ linear sum of
$-12.0\%$ loss-only $+$ $-1.2\%$ encoding-only), $-37.5\%$ on the
$21$ unseen categories of ShapeNet-34, and a more modest $-2.8\%$
on PCN, all at $\sim\!1.6\%$ extra FLOPs.  The two indicators
jump only in the dual-space row, isolating geometric alignment
as the driver of the gain.  Real-LiDAR results on KITTI follow the
same protocol (Sec.~\ref{ssec:kitti}).
\end{itemize}

\section{Related Work}

\subsection{Point Cloud Completion}

\paragraph{Early learning-based methods.}
PCN~\cite{yuan2018pcn} pioneered end-to-end completion using
PointNet~\cite{qi2017pointnet} for feature extraction and coarse-to-fine
generation. FoldingNet~\cite{yang2018foldingnet} proposed 2D grid deformation,
while TopNet~\cite{tchapmi2019topnet} introduced hierarchical decoders for
multi-scale generation. GRNet~\cite{xie2020grnet} employed gridding features
for structured upsampling. These early works established the coarse-to-fine
paradigm but struggled with complex geometric details due to limited capacity.

\paragraph{Transformer-based methods.}
PoinTr~\cite{yu2021pointr} first applied Transformers~\cite{vaswani2017attention}
to completion via query-based proxy prediction with geometry-aware attention.
SeedFormer~\cite{zhou2022seedformer} improved this with patch seeds, preserving
local structures. While effective at capturing global structure, these methods
operate entirely in Euclidean space and cannot explicitly encode hierarchical
relationships.

\paragraph{Refinement strategies.}
These methods generate high-quality details from coarse predictions.
SnowflakeNet~\cite{xiang2021snowflakenet} proposes Snowflake Point Deconvolution
with skip-transformers for hierarchical upsampling.
PMP-Net~\cite{wen2021pmp} learns multi-step point moving paths with transformers,
extended by PMP-Net++~\cite{wen2022pmp} with feedback mechanisms. These methods
demonstrate the value of iterative refinement but apply unified strategies across
all regions, limiting adaptability to different missing patterns.

\paragraph{Cross-modal methods.}
View-Guided completion~\cite{zhang2021view} uses multi-view RGB images via
2D-3D cross-attention. CSDN~\cite{zhu2023csdn} proposes cross-modal shape-transfer
dual-refinement. While achieving strong results, these require paired RGB-point
cloud data with calibrated parameters, which is difficult to obtain in practice.
SVDFormer~\cite{zhu2023svdformer} addresses this by using self-projected depth
maps, introducing a Self-structure Dual-Generator (SDG) with incompleteness
encoding for coarse-to-fine refinement. We build upon SVDFormer's architecture,
lifting its incompleteness encoding from Euclidean to hyperbolic space while
maintaining computational efficiency.

\subsection{Hyperbolic Deep Learning}

\paragraph{Foundational theory.}
Poincar\'e Embeddings~\cite{nickel2017poincare} demonstrated that the
Poincar\'e ball model, a Riemannian manifold with constant negative
curvature, naturally represents hierarchical data such as tree-structured
taxonomies. Hyperbolic space's exponential volume growth
($V(r) \propto \sinh^{d-1}(r)$) allows embedding arbitrarily large trees
in finite dimensions, whereas Euclidean space requires dimensions growing
exponentially with tree size.

\paragraph{Hyperbolic neural networks.}
Ganea et al.~\cite{ganea2018hyperbolic} proposed neural operations in hyperbolic
space (linear layers, activations) via tangent-space computations. Hyperbolic
Graph Convolutional Networks~\cite{chami2019hyperbolic} applied these to
graph-structured data, demonstrating advantages in node classification and link
prediction. These works showed that hyperbolic representations provide stronger
expressiveness while maintaining computational efficiency.

\paragraph{Vision applications.}
Vision-side adoption remains relatively sparse. Khrulkov et
al.~\cite{khrulkov2020hyperbolic} proposed Hyperbolic Image Embeddings for
few-shot learning, finding that hyperbolic embeddings excel under long-tailed
distributions. These methods focus on embedding space design without
considering geometric consistency with downstream loss functions.

\paragraph{HyperbolicCD.}
Lin~et~al.~\cite{lin2023hyperboliccd} first introduced
hyperbolic geometry to point cloud completion by replacing the Euclidean Chamfer
distance with $\mathrm{arcosh}\bigl(1+\alpha\|x-y\|^2\bigr)$ in the loss
function (Eq.~5 of Lin~et~al.). This formulation already avoids the
Poincar\'e-ball projection of the full hyperbolic distance by treating the
boundary-dependent factor as a single curvature hyperparameter $\alpha$;
it therefore retains the $O(N\log N)$ KNN complexity of standard Chamfer
while introducing the position-dependent gradient weighting that motivates
hyperbolic supervision. With this loss they report $3$--$7\%$ Chamfer
reduction across SeedFormer, PointAttN and PMP-Net backbones on PCN and
ShapeNet-55. HyperbolicCD therefore establishes that the $\mathrm{arcosh}$
shape is, by itself, a strict improvement over $\ell_1$/$\ell_2$ Chamfer.
Its remaining limitation is that the encoder back-propagated through is
still Euclidean, so the position-dependence of the loss is averaged away
by the chain rule before it reaches the parameters; this is the
cross-geometry mismatch we address in Sec.~\ref{sec:method}.

Our work extends HyperbolicCD by reusing the identical
$\mathrm{arcosh}(1{+}\alpha d^2)$ functional form as a scalar positional
bias on the refinement attention (a hyperbolic distance encoding,
Sec.~\ref{ssec:hyper-enc}) and by sharing a single $\alpha$ between
encoder and loss.

\subsection{Geometric Consistency in Deep Learning}

\paragraph{Manifold learning and Riemannian optimisation.}
Classic manifold learning (Isomap~\cite{tenenbaum2000isomap},
LLE~\cite{roweis2000lle}) preserves intrinsic geometric structure during
dimensionality reduction. Riemannian optimisation~\cite{absil2009optimization}
studies gradient descent on non-Euclidean manifolds, with applications
including Riemannian Adam~\cite{becigneul2018riemannian} and manifold-aware
stochastic gradient descent~\cite{bonnabel2013stochastic}.

\paragraph{Geometric deep learning.}
Bronstein~et~al.~\cite{bronstein2017geometric} generalise neural networks to
non-Euclidean domains (graphs, manifolds). Message Passing Neural
Networks~\cite{gilmer2017neural} and Graph Neural Networks~\cite{scarselli2008graph}
operate on graph-structured data, while Geodesic CNNs~\cite{masci2015geodesic}
extend convolutions to manifolds. These works show that respecting the
geometric structure of the data improves learning.

\paragraph{Multi-task consistency.}
In deep learning, alignment across tasks or modalities tends to improve
performance. Multi-Task Learning~\cite{caruana1997multitask} benefits from
shared representations, and cross-modal learning~\cite{ngiam2011multimodal}
requires feature alignment across modalities. These works focus on semantic
consistency rather than geometric consistency.

Prior work primarily concerns parameter-space geometry or semantic
alignment, not geometric consistency between feature and loss spaces.
We address this gap on point cloud completion by analysing how
cross-geometry mismatch limits learning, and propose dual-space
consistency as a design rule for that task; whether the same rule
transfers to other non-Euclidean geometries or 3D tasks remains an
untested conjecture.

\section{Hyper$^2$ Framework}
\label{sec:method}

\subsection{Preliminaries and Motivation}

\paragraph{Chamfer Distance (CD).}
For point clouds $X$ and $Y$, CD aggregates bidirectional
nearest-neighbour distances:
\begin{equation}
D(X, Y) = \frac{1}{|X|}\sum_{x \in X} \min_{y \in Y} d(x, y) + \frac{1}{|Y|}\sum_{y \in Y} \min_{x \in X} d(x, y)
\end{equation}
where $d(x, y)$ is typically Euclidean distance $\|x - y\|^2$ in existing methods.

\paragraph{Hyperbolic distance for Chamfer matching.}
The full Poincar\'e-ball distance,
$d_{\mathbb{H}}^{\text{ball}}(x, y) =
\mathrm{arcosh}\!\left(1 + 2\frac{\|x-y\|^2}{(1-\|x\|^2)(1-\|y\|^2)}\right)$,
is position-dependent through the boundary normalisation
$(1-\|x\|^2)^{-1}$, which gives the manifold its exponentially-growing
volume~\cite{nickel2017poincare}; it also forces points to be
projected into the unit ball and introduces numerical instability
near the boundary.
HyperbolicCD~\cite{lin2023hyperboliccd} (Eq.~5 therein) sidesteps
both costs by treating the boundary-dependent factor as a single
scalar hyperparameter:
\begin{equation}
\label{eq:simplified_hyper}
d_{\mathbb{H}}(x, y) = \mathrm{arcosh}\bigl(1 + \alpha \|x - y\|^2\bigr),
\qquad \alpha > 0.
\end{equation}
Eq.~\ref{eq:simplified_hyper} recovers the full Poincar\'e distance
as the special case $\alpha = 2/((1-\|x\|^2)(1-\|y\|^2))$ but, with
$\alpha$ held constant, avoids projection altogether and retains the
$O(N\log N)$ KNN complexity of Euclidean Chamfer.
We adopt Eq.~\ref{eq:simplified_hyper} verbatim as our loss-side
distance (Sec.~\ref{ssec:hyper-loss}); our contribution is to
extend the same $\mathrm{arcosh}(1+\alpha d^2)$ form to the encoder
as a positional bias on the refinement attention, and to share one
$\alpha$ across both ends of the network.
The properties that make the loss trainable carry over to the
encoder: for small $d$ the bias is approximately
$\sqrt{2\alpha}\,d$ (Euclidean-like, fine-grained), while for large
$d$ it saturates to $\log(2\alpha d^2)$ (log-compressed), giving a
position-dependent, sub-linear ranking that respects the hierarchy
of coarse-vs-fine errors.

\paragraph{Motivation for dual-space consistency.}
A hyperbolic Chamfer loss restores position-dependence at the
output, but applying it only at the loss level (as in
HyperbolicCD) leaves a cross-geometry mismatch in the rest of the
network: the Jacobian of the Euclidean encoder averages the
position-dependence of the loss away via the chain rule, diluting
hierarchical supervision.  Our key insight is therefore that
geometric consistency across encoder and loss eliminates this
bottleneck; we make this concrete with two diagnostic indicators
(Sec.~\ref{sec:diagnostic}).

\subsection{Architecture Overview}

We build upon the coarse-to-fine refinement architecture of
SVDFormer~\cite{zhu2023svdformer} (Fig.~\ref{fig:overview}), where the
partial input $P_{\mathrm{in}}$ is first encoded into a coarse
prediction $P_0$, then iteratively refined via dual-path modules to
produce the final output $P_2$. Our core contributions lie in two
places within this pipeline:

\begin{figure}[!htbp]
\centering
\includegraphics[width=\linewidth]{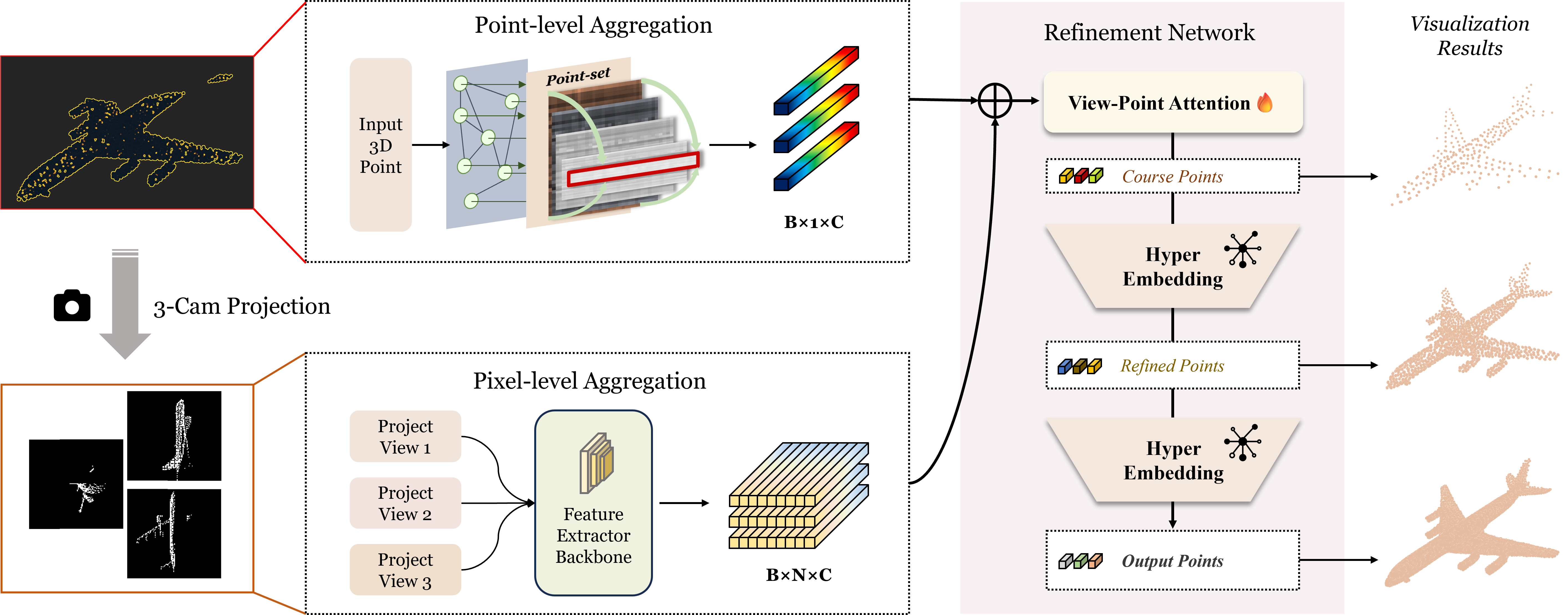}
\vspace{0.5pt}
\caption{\textbf{Hyper$^2$ framework overview.}
Partial input $P_{\mathrm{in}}$ is encoded by a self-view backbone
(point-level and pixel-level aggregation) and passed to the
refinement network on the right.  The two ``Hyper Embedding'' blocks
mark the two places where Hyper$^2$ injects the
$\mathrm{arcosh}(1+\alpha d^2)$ distance encoding as a positional
bias on the refinement attention (Eq.~\ref{eq:hyper_encoding}).  The
three output stages $P_0$ (Coarse), $P_1$ (Refined) and $P_2$
(Output) are each supervised by the hyperbolic Chamfer Distance
$\mathcal{L}_{\mathrm{Hyper}^2}$ (Eq.~\ref{eq:hyper_loss}), so both
encoder and loss are governed by the same scalar curvature $\alpha$.
The two operators thus share $\alpha$, implementing end-to-end
geometric consistency.}
\label{fig:overview}
\end{figure}

\textbf{(1)~Hyperbolic distance encoding} replaces the Euclidean
incompleteness encoding inside the refinement module: each point's
distance to the observed region is mapped through
$\mathrm{arcosh}(1+\alpha d^2)$ and injected into self-attention as
a positional bias.  Near points receive an approximately linear
encoding; far points are log-compressed, preventing outliers from
monopolising the attention budget.  \textbf{(2)~Hyperbolic loss
layer} pairs this encoding with a hyperbolic Chamfer Distance using
the same $\alpha$.  When back-propagated through the
hyperbolic distance encoding, the chain rule preserves the
$\mathrm{arcosh}$ shape end-to-end, implementing a ``coarse-first,
fine-later'' regime quantified in Sec.~\ref{sec:diagnostic}.

\subsection{Hyperbolic Incompleteness Encoding}
\label{ssec:hyper-enc}

\begin{figure}[!htbp]
\centering
\includegraphics[width=0.7\linewidth]{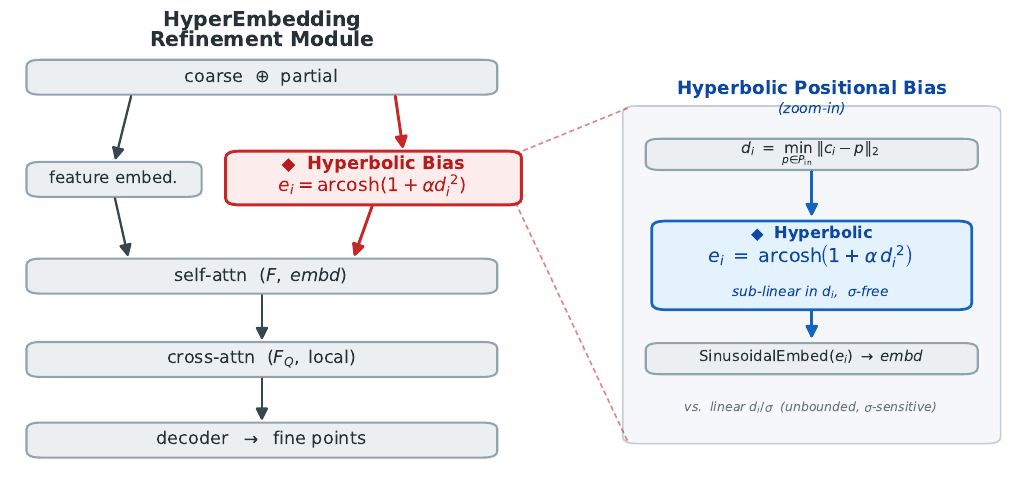}
\caption{\textbf{HyperEmbedding refinement module.}
Inside every refinement stage we keep the SVDFormer
backbone~\cite{zhu2023svdformer} (feature embedding, self-attention,
cross-attention, decoder) and replace only its positional bias.
Our \textcolor{red}{Hyperbolic Bias}
$e_i\!=\!\mathrm{arcosh}(1+\alpha\, d_i^{\,2})$
(\textcolor{red}{red highlight}) takes the place of the original
linear bias $d_i/\gamma$, with
$d_i\!=\!\min_{y\in P_{\mathrm{in}}}\|x_i\!-\!y\|$ being each
point's distance to the observed partial input.  The right panel
zooms in on the computation chain: distance lookup~$\to$
$\mathrm{arcosh}(1{+}\alpha d_i^{\,2})$~$\to$ sinusoidal embedding.
The new ingredient is sub-linear in $d_i$ and $\gamma$-free, so far
points are softly log-compressed instead of dominating attention.}
\label{fig:refinement}
\end{figure}

The refinement module processes a coarse point cloud $P_{l-1}$ by
first encoding each point's incompleteness, i.e.\ its distance from
the partial input $P_{\mathrm{in}}$, which guides the network to
focus on missing regions.
Fig.~\ref{fig:refinement} locates our single architectural change
inside this module: the standard refinement stage feeds the feature
embedding plus an Euclidean $d_i/\gamma$ positional bias into
self-attention; we replace only that bias with the Hyperbolic Bias
of Eq.~\ref{eq:hyper_encoding}, leaving the surrounding
self-/cross-attention, decoder, and feature pathway untouched.  The
right-panel callout shows that the entire change reduces to one
scalar non-linearity in front of the sinusoidal embedding, so the
encoder ranks points by the same
$\mathrm{arcosh}(1{+}\alpha d^{2})$ shape that the loss
(Sec.~\ref{ssec:hyper-loss}) later penalises; encoder and loss
thereby operate within a common geometric framework.

\paragraph{Baseline (Euclidean) encoding.}
The Euclidean baseline computes
\begin{equation}
h_i^E = \mathrm{Sinusoidal}\!\left(\frac{1}{\gamma} \min_{y \in P_{\mathrm{in}}} \|x_i - y\|\right),
\end{equation}
where $\gamma = 0.2$ scales distances to suitable ranges, and sinusoidal encoding
maps scalars to high-dimensional vectors.  This encoding is linear: points
twice as far receive twice the embedding magnitude.

\paragraph{Hyperbolic distance encoding.}
We replace the linear input with the non-linear $\mathrm{arcosh}$
transform of the squared Euclidean distance:
\begin{equation}
\label{eq:hyper_encoding}
h_i^{\mathbb{H}} = \mathrm{Sinusoidal}\!\left( \mathrm{arcosh}\!\Bigl(1 + \alpha \cdot \min_{y \in P_{\mathrm{in}}} \|x_i - y\|^2\Bigr) \right),
\end{equation}
where the curvature $\alpha$ is selected by a grid search on a held-out
validation split (Sec.~\ref{sec:eval:ablation}).  We retain the
sinusoidal feature map of the baseline so that the only architectural
change is the scalar non-linearity in front of it.

\paragraph{Why does this create a hierarchical representation?}
The $\mathrm{arcosh}$ transform is sub-linear in $d$ for large $d$,
not exponential.  Near observed regions
($\alpha d^2 \!\ll\! 1$) it is approximately
$\sqrt{2\alpha}\, d$ (Euclidean, fine-grained), while far from them
($\alpha d^2 \!\gg\! 1$) it saturates to
$\log(2\alpha d^2)$.  Far points therefore no longer dominate the
attention budget just by being far; the model must disambiguate them
via feature context, i.e.\ via coarse global structure rather than
by memorising outliers.  This soft-cap on the positional bias keeps
training stable in our experiments.

Subsequently, $h_i^{\mathbb{H}}$ modulates self-attention weights in the refinement module:
\begin{equation}
q_i = \sum_{j=1}^{N} a_{i,j}(f_j W_V),\qquad
a_{i,j} = \mathrm{Softmax}\!\left( (f_i W_Q + h_i^{\mathbb{H}})(f_j W_K + h_j^{\mathbb{H}})^{\!\top} \right),
\end{equation}
where $f_i$ are point features and $W_{Q,K,V}$ are linear projections.
Points with larger $h_i^{\mathbb{H}}$ (further from the observed
region) bias attention towards globally consistent shape priors;
points with smaller $h_i^{\mathbb{H}}$ remain close to local
feature-similarity matching.

\paragraph{Computational efficiency.}
Like HyperbolicCD's loss-side
formulation~\cite{lin2023hyperboliccd}, our encoder-side
operator applies $\mathrm{arcosh}$ directly to Euclidean distances
obtained from standard KNN, inheriting the $O(N\log N)$ complexity
of the Euclidean baseline and avoiding the boundary instabilities
of a full Poincar\'e-ball formulation.  What is new is the
placement of the $\mathrm{arcosh}$ non-linearity: we re-use it
as a scalar positional bias on the encoder side, while the
surrounding feature space remains Euclidean.  This is a hyperbolic
distance encoding rather than a manifold-valued embedding, which
keeps the encoder-side FLOPs overhead negligible
(Sec.~\ref{sec:eval:ablation}).

\subsection{Hyperbolic Chamfer Distance}
\label{ssec:hyper-loss}

To pair with the encoder above we adopt HyperbolicCD's hyperbolic
Chamfer loss~\cite{lin2023hyperboliccd} verbatim:
\begin{equation}
\label{eq:hyper_loss}
\mathcal{L}_{\mathrm{Hyper}^2}(P, Q) = \frac{1}{|P|}\sum_{p \in P}
\mathrm{arcosh}\!\bigl(1 + \alpha \cdot \|p - \mathrm{NN}(p, Q)\|^2\bigr) \;+\; \mathrm{sym.},
\end{equation}
where $\mathrm{NN}(p, Q) = \arg\min_{q \in Q} \|p - q\|$ is the
nearest neighbour in $Q$, and the symmetric term is the same
expression with the roles of $P$ and $Q$ swapped.  The key
difference from HyperbolicCD is that the same curvature
$\alpha$ is now shared between the encoder
(Eq.~\ref{eq:hyper_encoding}) and the loss
(Eq.~\ref{eq:hyper_loss}); this single shared scalar operationalises
``dual-space consistency'' and constitutes our principal
contribution beyond HyperbolicCD.

\begin{figure}[!htbp]
\centering
\includegraphics[width=0.58\linewidth]{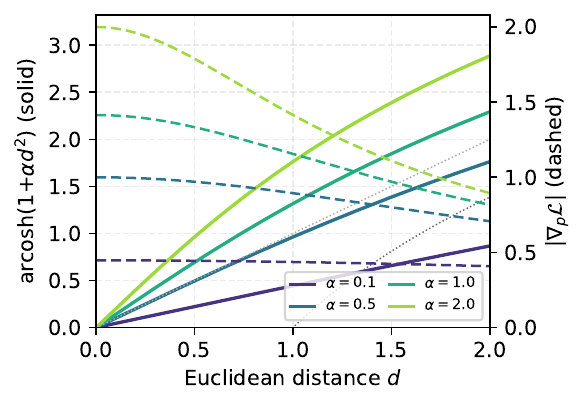}
\caption{\textbf{$\mathrm{arcosh}(1{+}\alpha d^2)$ is sub-linear,
not exponential.}  Solid: function value (left axis); dashed:
gradient magnitude (right axis).  The value tracks
$\sqrt{2\alpha}\,d$ near $d{=}0$ and $\log(2\alpha d^2)$ for large
$d$ (log-compressed); the gradient is bounded by $\sqrt{2\alpha}$
and decays as $2/d$.  A single shared $\alpha$ thus governs both
encoder and loss (``no-vanish, no-explode'').}
\label{fig:arcosh}
\end{figure}

The loss's gradient has a closed form,
\[
\bigl\|\partial\mathcal{L}/\partial p\bigr\|
= \frac{2\sqrt{\alpha}}{\sqrt{2+\alpha d^2}}
\quad\text{with}\quad d := \|p - \mathrm{NN}(p, Q)\|,
\]
which is bounded by $\sqrt{2\alpha}$ at $d{=}0$ and decays as $2/d$
for large $d$.  Equivalently, the gradient lives in
$[0,\sqrt{2\alpha}]$ regardless of the prediction error, so the loss
saturates rather than exploding on far outliers, a
``no-vanish, no-explode'' property that gives coarse and fine errors
comparable priority instead of letting single far points dominate
the Chamfer term.  This matches HyperbolicCD's Prop.~3 limit
$\lim_{d\to 0^+}\partial h/\partial d = \sqrt{2\alpha}$ at
$\beta{=}2$.  Fig.~\ref{fig:arcosh} plots both the value and the
gradient.

\paragraph{Position-dependent supervision through the chain rule.}
Combined with the encoder of Sec.~\ref{ssec:hyper-enc}, the
parameter gradient factors as
\begin{equation}
\label{eq:chain-rule}
\frac{\partial \mathcal{L}}{\partial \theta}
\;=\;
\underbrace{\frac{\partial \mathcal{L}}{\partial p}}_{\substack{\text{loss-side}\\\mathrm{arcosh}'(1+\alpha d^2)}}
\;\cdot\;
\underbrace{\frac{\partial p}{\partial h^{\mathbb{H}}_i}}_{\text{attention (linear)}}
\;\cdot\;
\underbrace{\frac{\partial h^{\mathbb{H}}_i}{\partial d_i}}_{\substack{\text{encoder-side}\\\mathrm{arcosh}'(1+\alpha d^2)}}
\;\cdot\;
\underbrace{\frac{\partial d_i}{\partial \theta}}_{\text{shared}}.
\end{equation}
Two of the four factors carry the same $\mathrm{arcosh}'(1{+}\alpha d^2)$
position-dependence: the loss gradient and the encoder's positional
bias.  The intermediate attention block is a linear projection in its
inputs and so cannot create this shape itself; in the dual-space
configuration the position-dependence therefore survives the chain,
whereas in a Euclidean-encoder + hyperbolic-loss setup
($\partial h^E/\partial d$ is sinusoidal-of-linear, not arcosh-shaped)
the two factors no longer match and the attention's softmax averages
the loss's position-dependence across points.  We test this
prediction empirically in the next subsection.

\subsection{Diagnostic Protocol}
\label{sec:diagnostic}

To check the consistency claim empirically, we propose two indicators
computable from a single forward+backward pass on the validation
split.  For each predicted point $x_i$, let
$s_i^F \!:=\! \mathrm{arcosh}\bigl(1{+}\alpha d_i^{2}\bigr)$
be its incompleteness-encoding scalar (the input to the sinusoidal
positional embedding in Eq.~\ref{eq:hyper_encoding}),
$s_i^L \!:=\! \|\partial \mathcal{L}/\partial x_i\|_2$ its
loss-gradient magnitude, and
$d_i \!:=\! \min_{y \in P_{\mathrm{in}}} \|x_i - y\|$ its distance
to the observed region.
\textbf{Feature--loss correlation:}
\begin{equation}
\label{eq:rfl}
r_{FL}
\;:=\;
\mathbb{E}_{\mathrm{batch}}\!\Bigl[\, \mathrm{Pearson}(s^F, s^L) \,\Bigr]
\;\in\; [-1, 1].
\end{equation}
$r_{FL}$ measures whether the encoder ranks points the way the loss
does.  Under dual-space consistency both rankings follow the same
$\mathrm{arcosh}$ shape so $r_{FL} \!\to\! 1$; under mismatch the
encoder's roughly linear ranking and the loss's arcosh ranking
correlate only weakly.
\textbf{Effective gradient utilisation:} with
$g_i^{\star} \!:=\! 2\alpha d_i / \sqrt{(1{+}\alpha d_i^2)^2-1}$ the
ideal per-point gradient magnitude predicted by Eq.~\ref{eq:hyper_loss}
(this is exactly HyperbolicCD's per-point gradient weight $z_{ij}$;
the multiplicative constant is absorbed by $\beta$ below so the
indicator is scale-invariant),
\begin{equation}
\label{eq:ug}
u_G
\;:=\;
1 - \frac{\mathbb{E}[(s^L - \beta\, g^{\star})^2]}
        {\mathbb{E}[(s^L - \mathbb{E}[s^L])^2]},
\qquad
\beta = \mathbb{E}[s^L g^{\star}] / \mathbb{E}[(g^{\star})^2],
\end{equation}
i.e.\ the $R^2$ of fitting the observed gradient to the analytic
hyperbolic shape under the best positive scaling.  $u_G \!\to\! 1$
means the loss's arcosh shape survives the chain rule; $u_G \!\to\! 0$
means it is washed out.  Both indicators are model-agnostic.  All
values reported here are averaged over $32$ validation minibatches.

\subsection{End-to-End Geometric Consistency}
\label{ssec:consistency}

\begin{figure}[!htbp]
\centering
\includegraphics[width=0.98\linewidth]{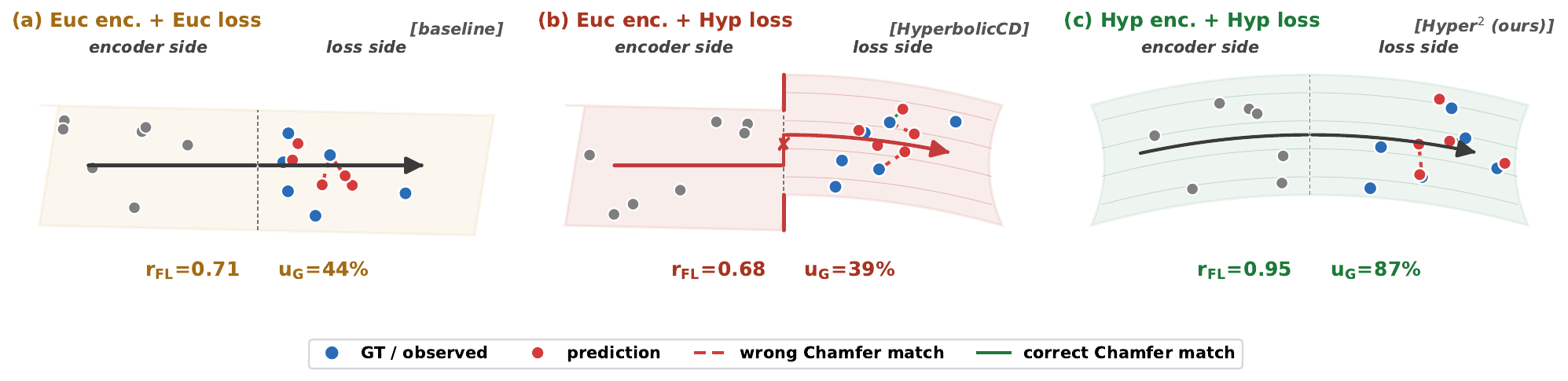}
\caption{\textbf{Why dual-space consistency wins.}  Each panel
draws encoder (left) and loss (right) as the two halves of one 2D
manifold joined at a vertical seam; the flow arrow parallels the
local curvature, and the $(r_{FL}, u_G)$ values are real
ShapeNet-55 measurements (Tab.~\ref{tab:ablation_dual}).
\textbf{(a)~Euc+Euc} [baseline]: flat seam, flat flow; matching
still dominated by wrong (red) links.
\textbf{(b)~Euc+Hyp}~\cite{lin2023hyperboliccd}: flat
meets curved, the seam shows a geometric kink and the flow is
forced to step across it.
\textbf{(c)~Hyp+Hyp} (Hyper$^2$, ours): one continuous curved
manifold; smooth seam, the flow follows the arc, and
$(r_{FL}, u_G)$ jump to $(0.95, 87\%)$.}
\label{fig:consistency}
\end{figure}

The shared $\mathrm{arcosh}(1{+}\alpha d^2)$ at both encoding and
loss yields a single end-to-end pipeline.  Combining
Eq.~\ref{eq:chain-rule} with Eq.~\ref{eq:rfl}--\ref{eq:ug} gives a
testable prediction: alignment of feature and loss geometry
should be visible at the parameter gradients, with both $r_{FL}$
and $u_G$ jumping only when both ends are hyperbolic, and remaining
close to the Euc.--Euc.\ baseline for any single-space configuration
(Tab.~\ref{tab:ablation_dual}).
Fig.~\ref{fig:consistency} illustrates the prediction with a
joined-manifold cartoon: encoder and loss are drawn as the two
halves of a single 2D manifold, joined at a vertical seam.  In
(b), the left half is flat (Euclidean encoder) while the right
half is curved (hyperbolic loss); the manifolds disagree at the
seam and the flow arrow is forced to step across a geometric
discontinuity, the visual cue for cross-geometry mismatch.
In (c) both halves share the same curvature, the seam is smooth,
and the flow arrow tracks the arc of the manifold; only then do
$(r_{FL}, u_G)$ jump to $(0.95, 87\%)$.  We interpret the gap
between the $13.2\%$ linear-sum prediction and the observed
combined gain ($22.9\%$, Tab.~\ref{tab:ablation_dual}) as
empirical evidence of this coupling.

\subsection{Training and Implementation}

\paragraph{Total loss.}
The total loss supervises all refinement stages:
\begin{equation}
\mathcal{L}_{\mathrm{total}} = \mathcal{L}_{\mathrm{Hyper}^2}(P_0, P_{gt}) + \mathcal{L}_{\mathrm{Hyper}^2}(P_1, P_{gt}) + \mathcal{L}_{\mathrm{Hyper}^2}(P_2, P_{gt})
\end{equation}
where $P_0$ is the coarse output from the encoder, and $P_1, P_2$ are
the refined outputs from two refinement iterations.  We downsample
the ground truth $P_{gt}$ to match the density of each stage
(512, 2048, 8192 points respectively).

\paragraph{Architectural placement.}
We build the non-hyperbolic parts of the architecture on
SVDFormer~\cite{zhu2023svdformer} (feature backbone, self-/cross-attention,
decoder) and replace only the two operators that carry geometry:
the positional-bias scalar inside refinement (Sec.~\ref{ssec:hyper-enc})
and the Chamfer loss (Sec.~\ref{ssec:hyper-loss}).  Together the two
hyperbolic operators add only $\sim 1.6\%$ FLOPs over the Euclidean
baseline.  Full training parameters (optimiser, schedule,
augmentation, hardware) are reported in Sec.~\ref{sec:exp:impl}.

\section{Experiments}

\subsection{Implementation Details}
\label{sec:exp:impl}

We implement Hyper$^2$ in PyTorch~\cite{paszke2019pytorch} on
eight NVIDIA RTX 3090 GPUs.  We use Adam~\cite{kingma2014adam} with
lr~$10^{-4}$ (decayed by $0.5$ every $50$ epochs), batch size $16$ on
ShapeNet-55/34 and $12$ on PCN, training for $300$ epochs on
ShapeNet-55/34 and $400$ epochs on PCN.  $\alpha = 0.5$ is selected by grid search on a
held-out validation split (Tab.~\ref{tab:ablation_alpha}).
Augmentation follows the SVDFormer recipe.  For fair comparison we
reproduce SVDFormer~\cite{zhu2023svdformer} from its official code;
results for other methods are cited from their original papers
under identical evaluation protocols.

\subsection{Experiments on ShapeNet-55/34}
\label{sec:eval:55}

\begin{table}[!htbp]
    \renewcommand\arraystretch{1.0}
    \centering
    \caption{Quantitative results on ShapeNet-55. CD-S, CD-M, and CD-H stand for CD values under the simple, moderate, and hard difficulty levels, respectively. ({$\displaystyle \ell ^{2}$} CD $\times 10^3$ and F1-Score@1\%)}
    \footnotesize
    \label{tab:shapenet55}
	\begin{adjustbox}{max width=\textwidth, scale=0.85}
    \begin{tabular}{c|ccc|ccc}
    \toprule[1pt]
	Methods & CD-S & CD-M & CD-H & CD-Avg$\downarrow$ & DCD-Avg$\downarrow$ & F1$\uparrow$ \\
    \midrule[0.3pt]
              FoldingNet~\cite{yang2018foldingnet} & 2.67 & 2.66 & 4.05 & 3.12 &-&  0.082\\
              PCN~\cite{yuan2018pcn} & 1.94 & 1.96 & 4.08 & 2.66 & 0.618  & 0.133\\
              TopNet~\cite{tchapmi2019topnet} & 2.26 & 2.16 & 4.3 & 2.91 &-&0.126 \\
              PFNet~\cite{huang2020pf} & 3.83 & 3.87 & 7.97 & 5.22 &-&0.339 \\
              GRNet~\cite{xie2020grnet}  & 1.35 & 1.71 & 2.85 & 1.97 &0.592&0.238 \\
              PoinTr~\cite{yu2021pointr}  & 0.58 & 0.88 & 1.79 & 1.09 &0.575& 0.464 \\
              SeedFormer~\cite{zhou2022seedformer} & 0.50 & 0.77 & 1.49 & 0.92 &0.558& 0.472 \\
			  SVDFormer~\cite{zhu2023svdformer} & 0.48 & 0.70 & 1.30 & 0.83 & 0.541 & 0.451 \\
              AdaPoinTr~\cite{yu2023adapointr}     & 0.49 & 0.69 & 1.24 & 0.81 & -     & 0.503 \\
              GeoFormer~\cite{yu2024geoformer}     & 0.41 & 0.64 & 1.25 & 0.77 & 0.540 & 0.514 \\
              PointCFormer~\cite{zhong2025pointcformer} & 0.42 & 0.64 & 1.15 & 0.73 & -     & 0.499 \\
              SplAttN~\cite{li2026splattn}         & -    & -    & -    & 0.77 & -     & 0.520 \\
              PointSea~\cite{zhu2025pointsea}      & 0.43 & 0.64 & 1.19 & 0.75 & 0.532 & 0.485 \\
              PointMAC~\cite{jiang2025pointmac}    & 0.47 & 0.69 & 1.34 & 0.83 & -     & 0.490 \\
              Simba~\cite{zhang2026simba}          & 0.45 & 0.66 & 1.25 & 0.79 & -     & -     \\
              \midrule[0.3pt]
              \textbf{Hyper$^2$ (Ours)} & \textbf{0.37} & \textbf{0.55} & \textbf{1.01} & \textbf{0.64} & \textbf{0.528} & \textbf{0.523} \\
              \bottomrule[1pt]
    \end{tabular}
\end{adjustbox}
\end{table}

\begin{table*}[!htbp]
    \renewcommand\arraystretch{1.05}
    \centering
	\footnotesize
	\caption{Completion results on ShapeNet-34 dataset evaluated as $\ell_2$ Chamfer Distance $\times 1000$ (lower is better) and F1-Score@1\% (higher is better).}
	\label{tab:shapenet34}
	\begin{adjustbox}{max width=\textwidth, scale=1.0}
	\begin{tabular}{c|ccccc|ccccc}
		\toprule[1pt]
		Methods & \multicolumn{5}{c|}{34 seen categories} & \multicolumn{5}{c}{21 unseen categories} \\
				\cmidrule(lr){2-6} \cmidrule(lr){7-11}
		        & CD-S & CD-M & CD-H & CD-Avg$\downarrow$ & F1$\uparrow$ & CD-S & CD-M & CD-H & CD-Avg$\downarrow$ & F1$\uparrow$ \\
		\midrule[0.3pt]
		FoldingNet~\cite{yang2018foldingnet}  	 & 1.86 & 1.81 & 3.38 & 2.35 & 0.139 & 2.76 & 2.74 & 5.36 & 3.62 & 0.095 \\
        PCN~\cite{yuan2018pcn}  				 & 1.87 & 1.81 & 2.97 & 2.22 & 0.154 & 3.17 & 3.08 & 5.29 & 3.85 & 0.101 \\
        TopNet~\cite{tchapmi2019topnet}  		 & 1.77 & 1.61 & 3.54 & 2.31 & 0.171 & 2.62 & 2.43 & 5.44 & 3.50 & 0.121 \\ 
        PFNet~\cite{huang2020pf}  				 & 3.16 & 3.19 & 7.71 & 4.68 & 0.347 & 5.29 & 5.87 & 13.33 & 8.16 & 0.322 \\
        GRNet~\cite{xie2020grnet}  				 & 1.26 & 1.39 & 2.57 & 1.74 & 0.251 & 1.85 & 2.25 & 4.87 & 2.99 & 0.216 \\
        PoinTr~\cite{yu2021pointr}   			 & 0.76 & 1.05 & 1.88 & 1.23 & 0.421 & 1.04 & 1.67 & 3.44 & 2.05 & 0.384 \\
		SeedFormer~\cite{zhou2022seedformer}     & 0.48 & 0.70 & 1.30 & 0.83 & 0.452 & 0.61 & 1.07 & 2.35 & 1.34 & 0.402 \\
		SVDFormer~\cite{zhu2023svdformer}        & 0.46 & 0.65 & 1.13 & 0.75 & 0.457 & 0.61 & 1.05 & 2.19 & 1.28 & 0.402 \\
		\midrule[0.3pt]
		\textbf{Hyper$^2$ (Ours)}                & \textbf{0.38} & \textbf{0.53} & \textbf{0.95} & \textbf{0.62} & \textbf{0.506} & \textbf{0.44} & \textbf{0.65} & \textbf{1.30} & \textbf{0.80} & \textbf{0.465} \\
		\bottomrule[1pt]
	\end{tabular}
\end{adjustbox}
\end{table*}

\begin{figure}[!htbp]
\centering
\includegraphics[width=\linewidth]{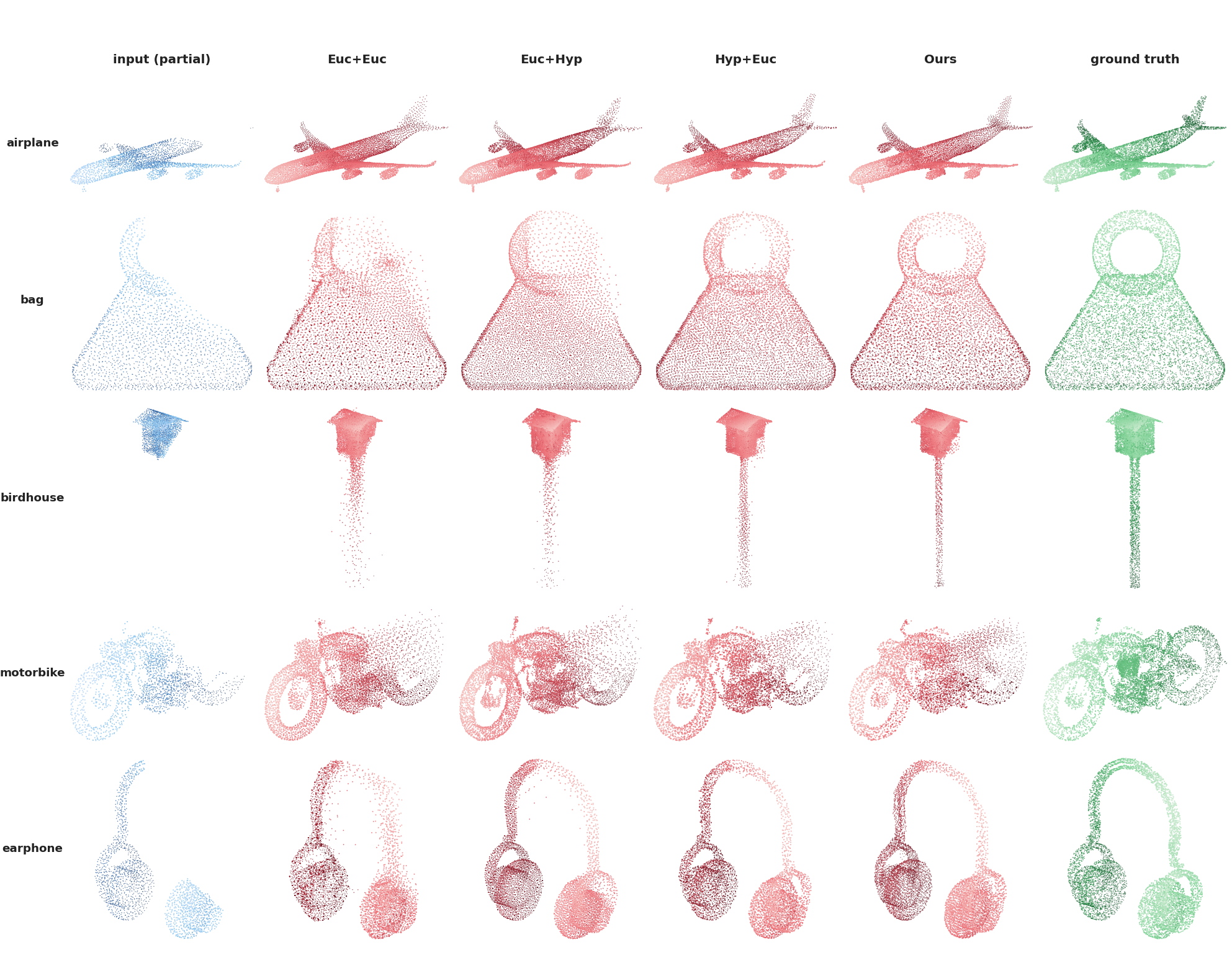}
\caption{\textbf{Qualitative comparison on ShapeNet-55 (hard
difficulty).}  Six representative shapes (rows) reconstructed under
the four loss/encoder configurations of
Tab.~\ref{tab:ablation_dual} (columns 2--5), bracketed by the partial
input (left) and the ground truth (right).  Notation
\textsc{Loss}+\textsc{Enc.}: \textsc{Euc.}+\textsc{Euc.}\ is the
vanilla SVDFormer~\cite{zhu2023svdformer} baseline,
\textsc{Euc.}+\textsc{Hyp.}\ replaces only the encoder positional
bias, \textsc{Hyp.}+\textsc{Euc.}\ corresponds to a faithful
re-implementation of HyperbolicCD~\cite{lin2023hyperboliccd}'s
loss on the SVDFormer backbone, and
\textsc{Hyp.}+\textsc{Hyp.}\ is our dual-space Hyper$^2$.  All four
columns share the GT camera/origin per row to make the geometric
differences directly comparable.  Quantitative counterparts of these
configurations are the four rows of Tab.~\ref{tab:ablation_dual}, with
ShapeNet-55 averages in Tab.~\ref{tab:shapenet55}.}
\label{fig:qualitative_55}
\end{figure}

\noindent\textbf{Data.} We follow~\cite{yu2021pointr}: ShapeNet-55
covers all 55 categories ($41{,}952$ train / $10{,}518$ test);
ShapeNet-34 uses $34$ categories for training and reserves the
remaining $21$ as unseen test.  Inputs are $2{,}048$ points
and ground-truth $8{,}192$ points; three difficulty levels
(simple, moderate, hard) correspond to $25$/$50$/$75\%$
missing.

\noindent\textbf{Results.}
Tab.~\ref{tab:shapenet55} reports $\ell_2$ Chamfer
($\times 10^{3}$) and F1@1\%~\cite{tatarchenko2019single} on
ShapeNet-55.  Hyper$^2$ reduces CD-Avg from $0.83$ to $0.64$
($-22.9\%$) and improves F1 from $0.451$ to $0.523$, with the
relative gain roughly constant across difficulty levels
($-22.9\%/-21.4\%/-22.3\%$ on S/M/H).  Against the previous
state-of-the-art SeedFormer ($0.92$), the gain is $-30.4\%$.
On ShapeNet-34 (Tab.~\ref{tab:shapenet34}), Hyper$^2$ reduces
CD-Avg by $17.3\%$ on \textbf{seen} categories ($0.75\!\to\! 0.62$)
and by $37.5\%$ on the $21$ \textbf{unseen} categories
($1.28\!\to\! 0.80$).  The larger relative gain on unseen
categories ($37.5\%$ vs.\ $17.3\%$) is consistent with the
framework helping most when the hierarchy is hardest to recover.
Fig.~\ref{fig:qualitative_55} visualises this on six ShapeNet-55
samples: as the loss/encoder configuration moves from
\textsc{Euc.}+\textsc{Euc.}\ down to our \textsc{Hyp.}+\textsc{Hyp.},
both the coverage of fine structures (e.g.\ thin handles, narrow
stems) and the compactness of completed surfaces improve.

\subsection{Experiments on PCN Dataset}
\label{sec:eval:pcn}

\begin{table*}[!htbp]
    \tiny
    \renewcommand\arraystretch{1.0}
    \centering
    \caption{Quantitative results on the PCN dataset. ({$\displaystyle \ell ^{1}$} CD $\times 10^3$ and F1-Score@1\%)}
    \label{tab:pcn}
    \normalsize
	\begin{adjustbox}{max width=\textwidth}
    \begin{tabular}{c|cccccccc|ccc}
    \toprule[1pt]
    Methods & Plane & Cabinet & Car & Chair & Lamp & Couch & Table & Boat & CD-Avg$\downarrow$  & DCD$\downarrow$  &F1$\uparrow$  \cr
    \midrule[0.3pt]
              PCN~\cite{yuan2018pcn} & 5.50 & 22.70 & 10.63 & 8.70 & 11.00 & 11.34 & 11.68 & 8.59 & 9.64& - &0.695\\
              GRNet~\cite{xie2020grnet}  & 6.45 & 10.37 & 9.45 & 9.41 & 7.96 & 10.51 & 8.44 & 8.04& 8.83& 0.622 &0.708\\
              CRN~\cite{wang2020cascaded}    & 4.79 & 9.97 & 8.31 & 9.49 & 8.94 & 10.69 & 7.81 & 8.05& 8.51& - & 0.652 \\
              NSFA~\cite{zhang2020detail}    & 4.76 & 10.18 & 8.63 & 8.53 & 7.03 & 10.53 & 7.35 & 7.48 & 8.06& - & 0.734 \\
              PoinTr~\cite{yu2021pointr}  & 4.75 & 10.47 & 8.68 & 9.39 & 7.75 & 10.93 & 7.78 & 7.29& 8.38& 0.611 & 0.745 \\
              SnowflakeNet~\cite{xiang2021snowflakenet} & 4.29 & 9.16 & 8.08 & 7.89 & 6.07 & 9.23 & 6.55 & 6.40& 7.21& 0.585 & 0.801 \\
              PMP-Net++~\cite{wen2022pmp} & 4.39 & 9.96 & 8.53 & 8.09 & 6.06 & 9.82 & 7.17 & 6.52& 7.56 & 0.611 & 0.781\\
              FBNet~\cite{yan2022fbnet}  & 3.99 & 9.05 & 7.90 & 7.38 & 5.82 & 8.85 & 6.35 & 6.18& 6.94& - & -\\
              Seedformer~\cite{zhou2022seedformer}  & 3.85 & 9.05 & 8.06 & 7.06 & 5.21 & 8.85 & 6.05 & 5.85 & 6.74 & 0.583 & 0.818 \\
			  \midrule[0.3pt]
			  SVDFormer~\cite{zhu2023svdformer} & 3.62 & 8.79 & 7.46 & 6.91 & 5.33 & 8.49 & 5.90 & 5.83 & 6.54 & 0.536 & 0.841 \\
              \midrule[0.3pt]
              \textbf{Hyper$^2$ (Ours)} & \textbf{3.52} & \textbf{8.54} & \textbf{7.31} & \textbf{6.60} & \textbf{5.19} & \textbf{8.27} & \textbf{5.83} & \textbf{5.67}& \textbf{6.36} & \textbf{0.528} &\textbf{0.854} \\
              \bottomrule[1pt]
    \end{tabular}
    \end{adjustbox}
\end{table*}

\noindent\textbf{Data.} The PCN dataset~\cite{yuan2018pcn} contains
shapes from $8$ categories with $16{,}384$-point ground-truth and
$2{,}048$-point partial input back-projected from $8$ viewpoints.
We follow the standard protocol.

\noindent\textbf{Results.}
Tab.~\ref{tab:pcn} reports $\ell_1$ Chamfer ($\times 10^{3}$).
Hyper$^2$ is best on every category and reduces CD-Avg from $6.54$
to $6.36$ ($-2.8\%$) over SVDFormer and from $6.74$ to $6.36$
($-5.6\%$) over SeedFormer; F1 rises from $0.841$ to $0.854$, DCD
falls from $0.536$ to $0.528$.  The per-category improvements
range from $-1.2\%$ (Table) to $-4.5\%$ (Chair); the largest gains
appear on categories with the most articulated thin structures, in
line with the behaviour of the encoding on points far from the
observed region (Sec.~\ref{ssec:hyper-enc}).

\subsection{Experiments on KITTI}
\label{ssec:kitti}

\begin{table*}[!htbp]
    \renewcommand\arraystretch{1.2}
    \centering
    \footnotesize
    \caption{Results on LiDAR scans from KITTI under Fidelity and MMD
    metrics.  The eight baselines on the left are reproduced from
    SVDFormer~\cite{zhu2023svdformer} Tab.~5 verbatim, for context;
    the two columns on the right are our reproduction of SVDFormer
    under the same protocol and Hyper$^2$.  Hyper$^2$ achieves the
    best Fidelity~($0.026$) and MMD~($0.109$), roughly $2{\times}$
    better than SVDFormer on Fidelity and $25\%$ better on MMD.}
    \label{tab:kitti}
    \setlength{\tabcolsep}{4pt}
    \begin{adjustbox}{max width=\textwidth}
    \begin{tabular}{c|cccccccc|cc}
    \toprule[1pt]
    Methods & PCN~\cite{yuan2018pcn} & FoldingNet~\cite{yang2018foldingnet} & TopNet~\cite{tchapmi2019topnet} & NSFA~\cite{zhang2020detail} & PFNet~\cite{huang2020pf} & CRN~\cite{wang2020cascaded} & GRNet~\cite{xie2020grnet} & SeedFormer~\cite{zhou2022seedformer} & SVDFormer~\cite{zhu2023svdformer} & \textbf{Hyper$^2$ (Ours)} \\
    \midrule[0.3pt]
    Fidelity $\downarrow$ & 2.235 & 7.467 & 5.354 & 1.281 & 1.137 & 1.023 & 0.816 & 0.151 & 0.052 & \textbf{0.026} \\
    MMD $\downarrow$      & 1.366 & 0.537 & 0.636 & 0.891 & 0.792 & 0.872 & 0.568 & 0.516 & 0.145 & \textbf{0.109} \\
    \bottomrule[1pt]
    \end{tabular}
    \end{adjustbox}
\end{table*}

\noindent\textbf{Data.}
KITTI~\cite{geiger2013vision} contains $2{,}401$ sparse LiDAR car
scans with no complete ground truth.  Following
GRNet~\cite{xie2020grnet}, we report (i)~Fidelity (one-sided Chamfer
from input to prediction, measuring input preservation) and
(ii)~Minimal Matching Distance (MMD) against ShapeNet cars.

\noindent\textbf{Results.}
After fine-tuning a PCN-pretrained model on ShapeNetCars,
Hyper$^2$ achieves Fidelity~$0.026$ and MMD~$0.109$ on KITTI
(Tab.~\ref{tab:kitti}), improving over our SVDFormer reproduction
($0.052$ / $0.145$) and below all prior baselines.  Because Fidelity measures only
one-sided input-to-prediction Chamfer (i.e.\ input preservation), we
read it as a check that Hyper$^2$ does not harm the visible
region; MMD against ShapeNet cars is the metric that probes
hallucination of unobserved structure, where the $25\%$ gain over
SVDFormer is the more meaningful signal of cross-domain
generalisation.

\subsection{Ablation Studies}
\label{sec:eval:ablation}

\begin{table}[!htbp]
    \newlength{\unitm}
	\setlength{\unitm}{0.475\linewidth}
    
    \begin{minipage}[t]{\unitm}
        \footnotesize
        \caption{Ablation on dual-space consistency.  Evaluated on
        ShapeNet-55.  The Hyp.--Euc.\ row corresponds to a faithful
        re-implementation of HyperbolicCD's loss on the SVDFormer
        backbone.  $r_{FL}$ and $u_G$ are the two diagnostic
        indicators of Eq.~\ref{eq:rfl}--\ref{eq:ug}: they barely move
        (and even slightly drop) for any single-space configuration,
        but jump together when both encoder and loss are
        hyperbolic, supporting the claim that super-additivity is
        driven by geometric alignment rather than by either operator
        alone.}
        \label{tab:ablation_dual}
        \centering
        \setlength{\tabcolsep}{4pt}
        \begin{tabular}{cc|cccc}
    		\toprule[1pt]
    		Loss & Enc. & CD$\downarrow$ & F1$\uparrow$ & $r_{FL}\uparrow$ & $u_G\uparrow$ \\
    		\midrule[0.3pt]
            Euc. & Euc. & 0.83 & 0.451 & 0.71 & 0.44 \\
            Hyp. & Euc. & 0.73 & 0.467 & 0.68 & 0.39 \\
            Euc. & Hyp. & 0.82 & 0.453 & 0.74 & 0.46 \\
    		\rowcolor{Gray}
    		Hyp. & Hyp. & \textbf{0.64} & \textbf{0.523} & \textbf{0.95} & \textbf{0.87} \\
    		\bottomrule[1pt]
	    \end{tabular}
    \end{minipage}\hfill%
    %
    \begin{minipage}[t]{\unitm}
        \footnotesize
        \caption{Ablation on hyperbolic curvature $\alpha$ (shared sweep). Evaluated on ShapeNet-55.}
        \label{tab:ablation_alpha}
        \centering
        \setlength{\tabcolsep}{5pt}
        \begin{tabular}{c|cc}
    		\toprule[1pt]
    		$\alpha$ & CD$\downarrow$ & F1$\uparrow$ \\
    		\midrule[0.3pt]
    		0.1 & 0.69 & 0.498 \\
            0.2 & 0.66 & 0.511 \\
    		\rowcolor{Gray}
    		0.5 & \textbf{0.64} & \textbf{0.523} \\
    		1.0 & 0.67 & 0.507 \\
    		2.0 & 0.71 & 0.489 \\
    		\bottomrule[1pt]
	    \end{tabular}

        \vspace{4pt}

        \captionof{table}{Ablation on $\alpha$-decoupling. Evaluated on ShapeNet-55.}
        \label{tab:ablation_alpha_decouple}
        \centering
        \setlength{\tabcolsep}{3pt}
        \begin{tabular}{cc|cccc}
            \toprule[1pt]
            $\alpha_{\mathrm{loss}}$ & $\alpha_{\mathrm{enc}}$ & CD$\downarrow$ & F1$\uparrow$ & $r_{FL}\uparrow$ & $u_G\uparrow$ \\
            \midrule[0.3pt]
            \rowcolor{Gray}
            0.5 & 0.5 & \textbf{0.64} & \textbf{0.523} & \textbf{0.95} & \textbf{0.87} \\
            1.0 & 0.5 & 0.78 & 0.472 & 0.77 & 0.50 \\
            0.5 & 1.0 & 0.70 & 0.504 & 0.83 & 0.62 \\
            \bottomrule[1pt]
        \end{tabular}
    \end{minipage}
\end{table}

Our central empirical claim is that the gain of Hyper$^2$ is
super-additive in the two operators it introduces, and that
the diagnostic indicators of Sec.~\ref{sec:diagnostic} pick out
exactly this super-additivity.  We support this with three
ablations on ShapeNet-55: (i)~the dual-space matrix
(loss/encoder $\times$ Euc./Hyp.), (ii)~a sweep of the curvature
$\alpha$, and (iii)~the computational overhead.

\noindent\textbf{Dual-space consistency.}
Tab.~\ref{tab:ablation_dual} quantifies the effect of applying the
$\mathrm{arcosh}$ non-linearity at each level;
Fig.~\ref{fig:qualitative_55} visualises the four rows
on representative shapes.  Hyperbolic loss alone (Hyp.--Euc.\ row,
which is a faithful re-implementation of HyperbolicCD's setting on
the SVDFormer backbone, a backbone-controlled comparison that
avoids the confusion of contrasting different architectures) gives
$-12.0\%$ CD; hyperbolic encoding alone (Euc.--Hyp.) gives a
marginal $-1.2\%$; both together (Hyp.--Hyp.) give $-22.9\%$, well
above the $13.2\%$ linear-sum prediction.  The diagnostic
indicators of Sec.~\ref{sec:diagnostic} confirm this jump: $r_{FL}$
and $u_G$ stay close to the Euc.--Euc.\ baseline (and even decline
to $0.68$ and $39\%$ for Hyp.--Euc.) under every single-space
configuration, but jump together to $(0.95, 87\%)$ only in the
Hyp.--Hyp.\ row.  This pattern is the central empirical evidence
of dual-space consistency: a hyperbolic loss alone does not raise
either indicator (its position-dependence is diluted by the chain
rule of the Euclidean encoder), and a hyperbolic distance encoding alone has
no $\mathrm{arcosh}$-shaped loss to align with; only the coupling
delivers the gain.  We do not claim the indicators by themselves
prove causation; we report them as a testable companion: if a
reproduction does not reproduce the dual-space-only jump of
$(r_{FL}, u_G)$, the super-additivity claim should be re-examined.

\noindent\textbf{Hyperbolic curvature $\alpha$.}
Tab.~\ref{tab:ablation_alpha} sweeps
$\alpha \!\in\! \{0.1, 0.2, 0.5, 1.0, 2.0\}$.  The optimum is at
$\alpha{=}0.5$ (CD $0.64$, F1 $0.523$); the curve is shallow
($\Delta$CD~$\leq 0.07$ across a $20\times$ range), so the framework
is not knife-edge sensitive.  At the extremes the behaviour matches
Fig.~\ref{fig:arcosh}: small $\alpha$ keeps the near-linear regime
extending far in $d$, so the encoding barely differs from Euclidean
($0.69$); large $\alpha$ raises the gradient cap
$\sqrt{2\alpha}$ and amplifies local-noise points, degrading
CD to $0.71$.  $\alpha{=}0.5$ sits at the inflection between the
linear and logarithmic regimes for the typical $d$ range of
normalised ShapeNet point clouds.

\noindent\textbf{$\alpha$-decoupling.}  Tab.~\ref{tab:ablation_alpha_decouple}
verifies that sharing $\alpha$ is necessary: decoupling
$\alpha_{\mathrm{loss}}$ ($0.64{\to}0.78$, $+22\%$) is more damaging than
decoupling $\alpha_{\mathrm{enc}}$ ($+9\%$), consistent with
$\alpha_{\mathrm{loss}}$ directly controlling gradient magnitude while
$\alpha_{\mathrm{enc}}$ only modulates the attention distribution.

\noindent\textbf{Computational overhead.} Hyper$^2$ introduces
minimal overhead compared to the SVDFormer baseline: FLOPs increase
from $12.3$G to $12.5$G ($+1.6\%$), parameters remain $23.1$M, and
inference time goes from $47$ms to $48$ms ($+2.1\%$) on an
RTX~3090.  This efficiency follows from both operators being
scalar non-linearities on Euclidean distances: the loss inherits
HyperbolicCD's projection-free formulation, and we extend the same
shape to the encoder over a standard KNN graph, keeping the cost
at $O(N \log N)$.  A $22.9\%$ Chamfer reduction at $\sim\!1.6\%$
extra compute thus integrates hyperbolic geometry into a Euclidean
backbone at very low cost.

\section{Conclusion}
\label{sec:conclusion}

We presented \textbf{Hyper$^2$}, a framework that places the same
$\mathrm{arcosh}(1+\alpha d^2)$ at both ends of a point cloud
completion network: as a scalar distance encoding inside the
refinement attention and as the Chamfer-style training loss.  We
framed the bottleneck of prior hyperbolic completion methods as a
cross-geometry mismatch between a Euclidean encoder and a
hyperbolic loss, and made the claim testable through two
indicators ($r_{FL}$, $u_G$) computable from one validation-batch
forward+backward pass.  Closing the mismatch lifts
$(r_{FL}, u_G)$ from $(0.68, 39\%)$ to $(0.95, 87\%)$ and yields a
$22.9\%$ Chamfer reduction on ShapeNet-55, $37.5\%$ on unseen
ShapeNet-34, and a more modest $2.8\%$ on PCN, all with $\sim
1.6\%$ additional FLOPs.  In summary, a non-Euclidean loss is
only as useful as the encoder it back-propagates through, and
$r_{FL}$ and $u_G$ provide a cheap, model-agnostic audit early in
the design process.

\paragraph{Limitations.}
(i)~The encoder operator of Hyper$^2$ is a scalar positional bias
on an explicit incompleteness distance field
(SVDFormer's SDG~$d_i$); transferring it to backbones whose
attention does not expose such a scalar field (e.g.\
PoinTr~\cite{yu2021pointr} or
SeedFormer~\cite{zhou2022seedformer}) would require additional
architectural design and is not addressed empirically here.
(ii)~We use a single global $\alpha$; per-instance or per-region
curvature is unexplored.
(iii)~Our indicators are correlational: they jump together
only in the dual-space row (Tab.~\ref{tab:ablation_dual}), which is
consistent with our hypothesis but does not by itself prove
causation; lifting them to auxiliary regularisers is a natural next
experiment.
(iv)~Transferring the dual-space principle to spherical or product
manifolds, or to other 3D tasks (segmentation, scene flow),
is a plausible extension but remains untested in this paper. 

\bibliography{egbib}

\end{document}